\documentclass[10pt,twocolumn,letterpaper]{article}

\usepackage{cvpr}              %

\definecolor{cvprblue}{rgb}{0.21,0.49,0.74}
\usepackage[pagebackref,breaklinks,colorlinks,allcolors=cvprblue]{hyperref}
\usepackage{graphicx}
\usepackage{booktabs}
\usepackage{multirow}
\usepackage{stackengine}
\usepackage{makecell}
\usepackage{wrapfig}
\usepackage{colortbl}
\usepackage{dashrule}
\usepackage[most]{tcolorbox}
\usepackage{algorithm}
\usepackage{algpseudocode}
\newcommand{\method}{ReMix}
\definecolor{light-gray}{gray}{0.6}
\colorlet{ourred}{orange!5}
\colorlet{ourblue}{blue!100}
\def\paperID{37521} %
\def\confName{CVPR}
\def\confYear{2026}

\title{Rejection Mixing: Fast Semantic Propagation of Mask Tokens\\for Efficient DLLM Inference}

\author{Yushi Ye$^{1,*}$, Feng Hong$^{1,*}$, Huangjie Zheng, Xu Chen$^1$\\
Zhiyong Chen$^1$, Yanfeng Wang$^2$, Jiangchao Yao$^{1,\dagger}$ \\ 
$^{1}$Cooperative Medianet Innovation Center, Shanghai Jiao Tong University\quad \\
$^{2}$School of Artificial Intelligence, Shanghai Jiao Tong University\\
{\tt\small \{stephen-ye, Sunarker\}@sjtu.edu.cn} 
}

\begin{document}
\maketitle
\begingroup
\renewcommand\thefootnote{}\footnotetext{$^*$Equal contribution. $^\dagger$Corresponding author.}
\endgroup
\begin{abstract}
Diffusion Large Language Models (DLLMs) promise fast non-autoregressive inference but suffer a severe quality-speed trade-off in parallel decoding. This stems from the ``combinatorial contradiction'' phenomenon, where parallel tokens form semantically inconsistent combinations. We address this by integrating continuous representations into the discrete decoding process, as they preserve rich inter-position dependency. 
We propose {\method} (Rejection Mixing), a framework that introduces a novel {Continuous Mixing State} as an intermediate between the {initial masked state} and the {final decoded token state}. This intermediate state allows a token's representation to be iteratively refined in a continuous space, resolving mutual conflicts with other tokens before collapsing into a final discrete sample. Furthermore, a rejection rule reverts uncertain representations from the continuous state back to the masked state for reprocessing, ensuring stability and preventing error propagation.
{\method} thus mitigates combinatorial contradictions by enabling continuous-space refinement during discrete diffusion decoding. Extensive experiments demonstrate that {\method}, as a training-free method, achieves a 2-8$\times$ inference speedup without any quality degradation.\footnote{Code: \href{https://github.com/Serpientw/ReMix-DLLM}{https://github.com/Serpientw/ReMix-DLLM}.}
\end{abstract}
    
\section{Introduction}\label{sec:intro}
The proliferation of Large Language Models (LLMs)~\citep{DBLP:journals/corr/abs-2303-18223,DBLP:journals/jmlr/ChowdheryNDBMRBCSGSSTMRBTSPRDHPBAI23,liu2023visual,peng2025hyperet,DBLP:journals/csur/CaoXYYYHZTOKS26} marks a pivotal moment in artificial intelligence. 
Yet their predominant autoregressive (AR) architecture, which generates outputs in a token-by-token sequence, severely limits inference speed.
To address this, Diffusion Large Language Models (DLLMs)~\citep{DBLP:journals/corr/abs-2508-10875,DBLP:journals/corr/abs-2502-09992,DBLP:journals/corr/abs-2508-15487,DBLP:journals/corr/abs-2506-20639} have emerged as a compelling non-AR paradigm. By modelling entire sequences of tokens in parallel, DLLMs bypass
the inherent dependencies of AR models, offering the potential for a substantial leap in generative throughput and establishing them as a critical frontier for developing high-efficiency language and multimodal systems~\citep{DBLP:journals/corr/abs-2505-15809,DBLP:journals/corr/abs-2505-16839,DBLP:journals/corr/abs-2505-16933}.

\begin{figure}[!t]\vspace{-10pt}
    \centering
    \includegraphics[width=\linewidth]{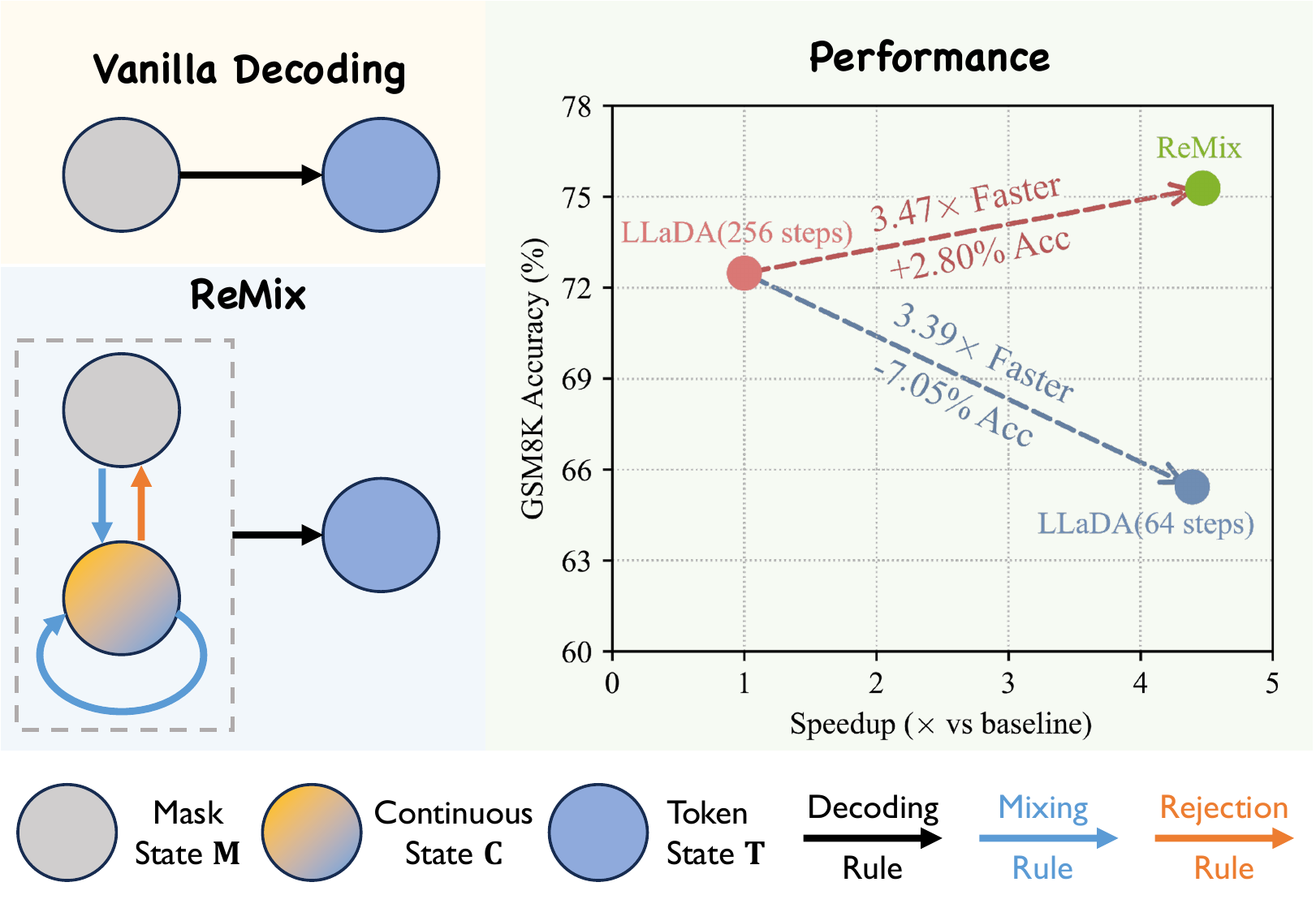}
    \vspace{-20pt}
    \caption{The {\method} decoding process and its impact on the quality-speed trade-off. (Left) By introducing an intermediate Continuous State ($\mathbf{C}$), {\method} extends the standard $\mathbf{M}\rightarrow T$ transition by introducing a flexible loop involving Mixing $\mathbf{M} \rightarrow \mathbf{C} $, self-transition $\mathbf{C} \circlearrowright$ and Rejection $\mathbf{C} \rightarrow \mathbf{M}$, enabling in-process refinement. (Right) On GSM8K, {\method} demonstrates superior performance, turning the baseline's 7.05\% accuracy loss (when accelerating 3.39x) into a 2.80\% accuracy gain while achieving similar acceleration.}
    \vspace{-18pt}
    \label{fig: intro}
\end{figure}

This parallel promise, however, is undermined by a critical dilemma: a severe quality-speed trade-off.
They achieve optimal performance when decoding a single token at a time, but the quality degrades significantly when attempting to decode multiple tokens in parallel~\citep{DBLP:journals/corr/abs-2502-09992}. 
This trade-off is a direct symptom of a fundamental problem we term the ``combinatorial contradiction", where tokens sampled within the same decoding step can be mutually contradictory. The issue is exemplified by \citet{DBLP:journals/corr/abs-2503-07154}: ``The list of poker hands that consist of two English words are: \_ \_.'' While valid subsequent pairs include ``high card'', ``two pair'', ``full house'', or ``straight flush'', a direct parallel approach might independently select the most probable token for each position, resulting in a semantically incorrect combination like ``high house''. Although sophisticated parallel decoding schedules~\citep{DBLP:journals/corr/abs-2505-22618,DBLP:journals/corr/abs-2505-24857,DBLP:journals/corr/abs-2506-10848,DBLP:journals/corr/abs-2509-25188,DBLP:journals/corr/abs-2509-26488} have improved the trade-off between quality and speed, they do not fully resolve the core consistency problem. WINO~\citep{DBLP:journals/corr/abs-2507-18578} mitigates this issue through a revocable strategy, but at the cost of introducing an additional verification block and its associated computational overhead. APD~\citep{DBLP:journals/corr/abs-2506-00413}, on the other hand, addresses this challenge by defining a multiplicative mixture between the DLLM marginal probabilities and the joint probability of sequences under a small auxiliary AR model, a method that, while effective, requires additional computation and enforces a left-to-right generation order.

In existing DLLMs,
``combinatorial contradiction''
arises from the discrete decoding nature: tokens generated within the same step cannot perceive one another and are sampled independently
~\citep{DBLP:conf/iclr/LiuBNB25}. 
Continuous representations offer a way out of this limitation: 
they enable the model to capture dependencies across different positions in the sequence and preserve richer information throughout the decoding process~\citep{DBLP:journals/corr/abs-2510-01329,DBLP:journals/corr/abs-2510-03206,DBLP:journals/corr/abs-2510-22510}.
To address combinatorial contradiction, we propose a novel approach that introduces continuous representations into the decoding process. By retaining continuous representations of unfilled positions throughout the decoding process, rather than remasking them and discarding valuable information, we preserve the interdependencies between positions in the sequence. This allows the model to iteratively refine the continuous representations over multiple decoding steps, resolving mutual contradictions and ensuring combinatorial consistency. 

Building on this intuition,
we propose {\method} (Rejection Mixing), a method that integrates continuous states into the discrete diffusion decoding process to address the ``combinatorial contradiction'' (see \cref{fig: intro}). 
In the standard decoding process, each token position transitions from a [MASK] state to a semantic token. Traditional methods use predefined rules to control this transition. ReMix introduces an intermediate continuous state, allowing token representations to be refined before finalizing them as discrete tokens. To ensure stability, we also incorporate a rejection mechanism that reverts uncertain representations back to the [MASK] state for reprocessing, preventing early uncertainty from affecting the decoding process.
Our main contributions are summarized as follows:

\begin{itemize}
    \item We analyze a fundamental bottleneck in the parallel decoding of DLLMs, which we term the ``combinatorial contradiction''. We posit that this issue stems from the purely discrete nature of the decoding process and propose a novel approach to resolve it by integrating continuous representations into the discrete diffusion loop.
    
    \item We propose {\method} (Rejection Mixing), a novel, training-free decoding framework. {\method} introduces a Continuous Mixing State as an intermediate step, enabling iterative refinement and ensuring stability through a rejection mechanism. This approach allows for resolving mutual dependencies in a continuous space before discretization, improving the overall decoding process.
    
    \item We conduct extensive experiments on both language and multimodal generation benchmarks. Our results demonstrate that {\method} successfully breaks the severe quality-speed trade-off, achieving a 2-8$\times$ inference speedup over strong baselines without any quality degradation, and in several cases, even improving output quality.
\end{itemize}

\section{Related Work}
\subsection{Diffusion-based Large Language Model}
Diffusion models~\citep{DBLP:conf/icml/Sohl-DicksteinW15,DBLP:conf/iclr/0011SKKEP21,DBLP:conf/nips/HoJA20,peng2026towards}, initially gaining prominence for image generation~\citep{DBLP:conf/nips/DhariwalN21,DBLP:conf/cvpr/RombachBLEO22,DBLP:conf/icml/NicholDRSMMSC22}, have recently emerged as a competitive alternative to AR language models in text generation. This transition from continuous to discrete data domains necessitates new modeling approaches. The foundational work by \citet{DBLP:conf/icml/Sohl-DicksteinW15} first explored this direction, which was later rigorously formalized by \citet{DBLP:conf/nips/AustinJHTB21} in the D3PM framework. D3PM conceptualizes the diffusion forward process as a discrete-state Markov chain with a series of transition matrices, which is later extended by \citet{DBLP:conf/nips/CampbellBBRDD22} using the theory of continuous-time Markov chains (CTMCs).

A particularly influential thread of research, known as masked diffusion models (MDMs), builds upon the absorbing-state diffusion concept in D3PM. As MDMs have evolved, they have demonstrated compelling performance across scales, from compact architectures like SEDD~\citep{DBLP:conf/icml/LouME24}, MDLM~\citep{DBLP:conf/nips/SahooASGMCRK24} and RADD~\citep{DBLP:conf/iclr/OuNXZSLL25} to large-scale implementations such as LLaDA~\citep{DBLP:journals/corr/abs-2502-09992} and Dream~\citep{DBLP:journals/corr/abs-2508-15487}. Most recently, this line of work has been expanded to multimodal learning. MMaDA~\cite{DBLP:journals/corr/abs-2505-15809}, for instance, introduces a family of multimodal large diffusion models featuring a shared probabilistic formulation and a modality-agnostic architecture, underscoring the versatility of the diffusion framework.

\begin{figure*}
    \centering
    \includegraphics[width=\linewidth]{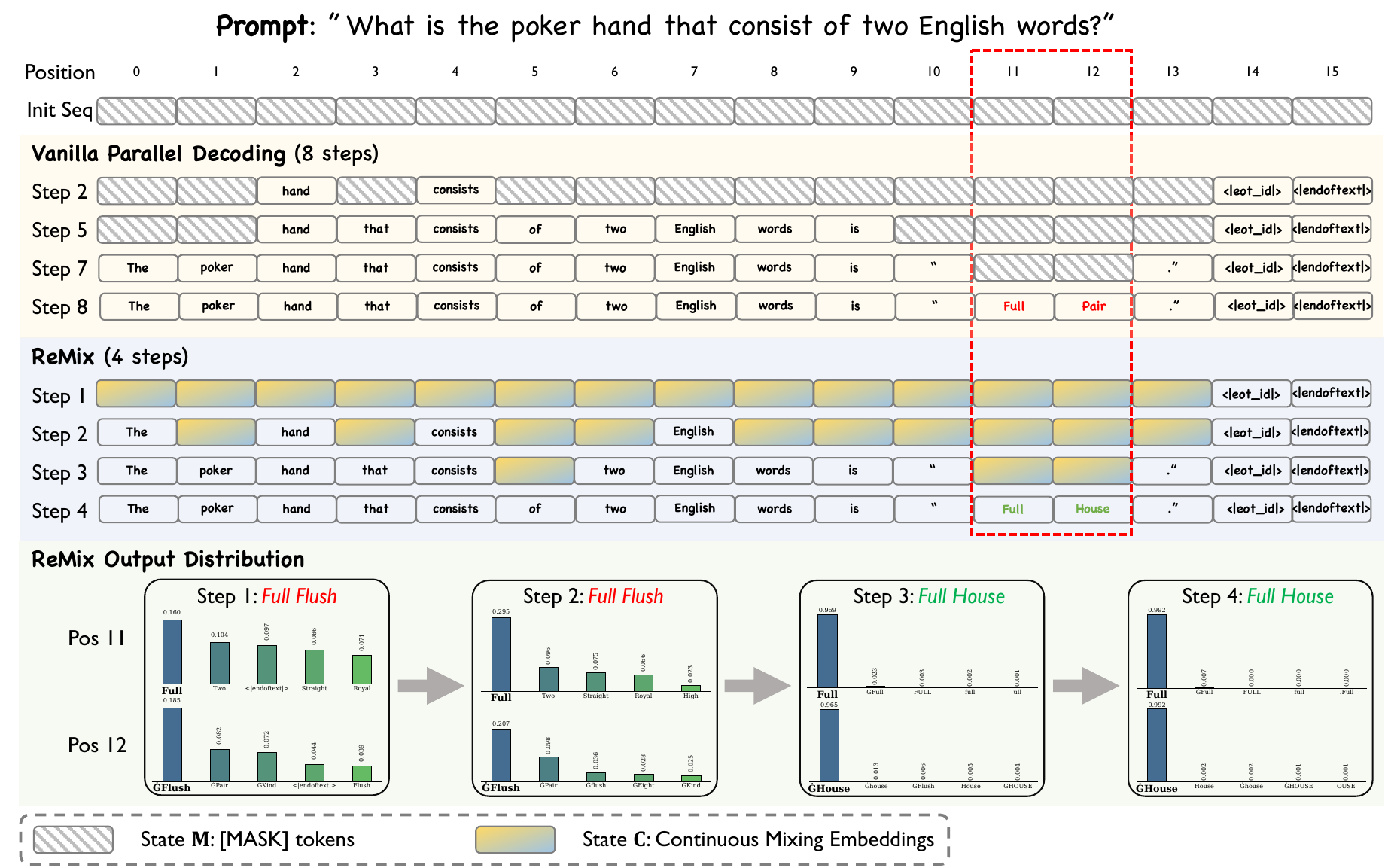}
    \caption{{\method} resolves combinatorial contradictions by allowing continuous-space refinement. Using the poker hand example, this figure compares the decoding processes. (Top) Vanilla Parallel Decoding: Succumbs to the contradiction. It independently decodes ``Full'' and ``Pair'' at positions 11 and 12, resulting in the semantically incorrect ``Full Pair" in 8 steps. (Middle) {\method}: Correctly generates ``Full House'' in only 4 steps. 
    (Bottom) {\method} Output Distribution: This visualizes the internal mechanism. In Steps 1 \& 2, both positions are in the Continuous Mixing State (State $\mathbf{C}$, gradient). Their output distributions are iteratively refined, allowing the model to recognize the ``Full'' $\leftrightarrow$ ``House'' dependency. This mutual coordination before discretization prevents the error and leads to the correct response. The \textnormal{\.{G}} symbol (\eg in \textnormal{\.{G}Flush} indicates a token that begins with a whitespace.}
    \vspace{-10pt}
    \label{fig: remix}
\end{figure*}

\subsection{Inference Acceleration Technique of DLLMs}
The acceleration of inference for DLLMs is an area of active research, as reducing latency and computational overhead is crucial for real-world applications. Broadly, two primary domains have emerged for accelerating inference: dynamic caching mechanism and optimized sampling strategy.

\noindent\textbf{Dynamic Caching Mechanism.} The caching mechanism for diffusion models differs from AR models~\citep{DBLP:conf/icml/JiangLZHHS0LLY25}, primarily due to the bidirectional attention mechanism employed in diffusion processes.
Several approaches have optimized caching in this context: 
Fast-dLLM~\citep{DBLP:journals/corr/abs-2505-22618} accelerates this by using chunked KV caching and confidence-aware parallel decoding; 
dKV-Cache~\citep{DBLP:journals/corr/abs-2505-15781} employs delayed caching to reuse key-value states after token decoding, ensuring stable acceleration; 
and Sparse-dLLM~\citep{DBLP:journals/corr/abs-2508-02558} reduces overhead by pruning less important tokens in the cache.

\noindent\textbf{Optimized Sampling Strategy.} Beyond dynamic caching, optimizing the sampling process itself represents a pivotal direction for accelerating inference in DLLMs. 
Research has explored diverse strategies: 
WINO~\citep{DBLP:journals/corr/abs-2507-18578} introduces revocable decoding to enhance both efficiency and effectiveness in DLLMs; 
similarly, FreeDave~\citep{DBLP:journals/corr/abs-2510-00294} enables lossless parallel decoding via self-drafting and self-verification;
COVER~\citep{DBLP:journals/corr/abs-2602-06161} performs context-preserving verification via KV-cache override to reduce flip-flop revisions in revocable diffusion decoding.
and Learn2PD~\citep{DBLP:journals/corr/abs-2509-25188} trains a lightweight filter model to predict token-wise convergence, reducing redundant decoding.

\section{Preliminary}\label{sec:preliminary}

\textbf{Decoding Process of DLLMs.} Let $\mathcal{V}$ be the vocabulary, in which $\text{[MASK]} \in \mathcal{V}$ is a special token. Given a prompt $X$, a DLLM parameterized with $\theta$ is designed to generate a response $Y = [y_1,y_2,\ldots,y_L]$ with a predefined response length $L$. The response sequence is initialized as all mask tokens, $Y\leftarrow [\text{[MASK]},\text{[MASK]},\ldots,\text{[MASK]}]$. The decoding process iteratively refines the response sequence $Y$
in each decoding step. We model the decoding process as a state transition, where each masked token (state $\mathbf{M}$) is transformed into a specific semantic token (state $\mathbf{T}$) under certain rules.

Specifically, at each decoding step, the model estimates the probability distribution over the response sequence as $p_{\theta}(\hat{Y}|X, Y)$. Tokens that meet the $\mathbf{M} \rightarrow \mathbf{T}$ transition condition are selected to be unmasked.
\begin{equation}
    y_i \leftarrow {\arg\max}_{v\in \mathcal{V}} p_\theta(\hat{y_i} = v | X, Y), \ \text{if} \ i \in \mathrm{rule}_{\mathbf{M} \rightarrow \mathbf{T}}. \nonumber
\end{equation}

Here, $\mathrm{rule}_{\mathbf{M} \rightarrow \mathbf{T}}$ denotes the set of positions that satisfy the $\mathbf{M} \rightarrow \mathbf{T}$ transition condition. 
For example, under the commonly used high-confidence greedy decoding, we have 
\begin{equation}
\mathrm{rule}_{\mathbf{M} \rightarrow \mathbf{T}} = \{ \mathop{\arg\max}_{i\in\{i|y_i=\text{[MASK]}\}} ( \max_{v \in \mathcal{V}} p_\theta(\hat{y_i} = v | X, Y)\}. \nonumber
\end{equation}

The decoding process concludes once every element in $Y$ has transitioned to state $\mathbf{T}$ as the final generated result.

\section{Method}\label{sec:method}
\subsection{Motivation} 
Current DLLMs still face the speed-performance trade-off, with optimal performance typically achieved when decoding one token at a time. Parallel decoding, while faster, often results in a performance drop due to ``combinatorial contradiction," where sampling multiple tokens from different plausible completion paths leads to mutual contradictions. We attribute this issue to the strictly \emph{discrete} nature of parallel decoding, where simultaneous sampling of multiple positions lacks coordination, causing semantic conflicts. To address this, we propose introducing a continuous intermediate state between the two discrete states, state $\mathbf{M}$ and state $\mathbf{T}$, allowing positions to accumulate and align decoding intent before finalizing the output, thereby resolving inconsistencies and improving performance (see \cref{fig: remix}).

\subsection{Rejection Mixing of Continuous States}
Building on the motivation above, we propose Rejection Mixing ({\method}), a method designed to accelerate semantic propagation in DLLM inference by introducing a continuous intermediate state $\mathbf{C}$ into the standard discrete $\mathbf{M} \rightarrow \mathbf{T}$  transition. This state allows for mutual perception and iterative adjustment of inter-positional inconsistencies before collapsing into a final discrete state T. 
Operationally, each position starts as a masked token in state $\mathbf{M}$. At each decoding step, if the decoding rule is satisfied, the position is directly decoded to the token state $\mathbf{T}$; otherwise, it shifts from $\mathbf{M}$ to the continuous state $\mathbf{C}$, where it is iteratively updated by the mixing rule, and either be accepted to $\mathbf{T}$ once the decoding rule holds or rejected back from $\mathbf{C}$ to $\mathbf{M}$ according to the rejection rule (detailed in ~\cref{alg:remix_corrected}).
We now detail the specific rules governing the transitions of each position among states $\mathbf{M}$, $\mathbf{C}$, and $\mathbf{T}$.

\noindent\textbf{Initialization ($\mathbf{M}$).} As introduced in \cref{sec:preliminary}, the decoding process begins with the entire response sequence initialized with mask tokens $\mathbf{M}$ at every position.
\begin{equation}
    y_i \leftarrow \text{[MASK]}, \ \forall i \in \{1,\ldots, L\}. \tag{Initialization}
\end{equation}

\noindent\textbf{Decoding Rule ($(\mathbf{M}, \mathbf{C}) \rightarrow \mathbf{T}$).}  For adaptive parallel decoding, we employ a commonly used confidence-based thresholding strategy. Specifically, when the confidence of the output distribution at a given position surpasses a predefined threshold $\tau_{\text{conf}}$, that position is decoded into a concrete semantic token, corresponding to a transition from state $\mathbf{M}$ (or $\mathbf{C}$) to state $\mathbf{T}$.
\begin{equation}
\begin{aligned}
    &y_i \leftarrow {\arg\max}_{v\in \mathcal{V}} \, p_{\theta}(\hat{y_i} = v | X, Y), \\
    &\text{if} \ {\max}_{v\in \mathcal{V}} p_{\theta}(\hat{y_i} = v | X, Y) > \tau_{\text{conf}}.
\end{aligned}
\tag{$(\mathbf{M}, \mathbf{C}) \rightarrow \mathbf{T}$}
\label{eq:decode}
\end{equation}

\noindent\textbf{Mixing Rule ($\mathbf{M} \rightarrow \mathbf{C} \circlearrowright$).} The Mixing Rule defines the transition from state $\mathbf{M}$ to the continuous state $\mathbf{C}$, as well as the self-transition process within $\mathbf{C}$. Since $\mathbf{C}$ is a continuous state rather than a discrete token, the transition to $\mathbf{C}$ is represented as an update to the embedding ${y}^{\text{emb}}$.
For discrete tokens, let the embedding layer be parameterized by a weight matrix $W \in \mathbb{R}^{|V| \times d}$, where $|V|$ is the vocabulary size and $d$ is the embedding dimension. Because the mapping between a token and its embedding is one-to-one, updating the token $y$ can be equivalently viewed as updating its embedding $y^{\text{emb}}$. Our Mixing Rule is designed to preserve the output distribution information from the previous step to enhance consistency in subsequent decoding.
\begin{equation}
\begin{aligned}
&y_i^{\text{emb}} \leftarrow \beta W^\top p_{\theta}(\hat{y_i} | X, Y) + (1-\beta) \mathrm{Emb}_{\text{[MASK]}},\\
& \text{if position $i$ not decoded}. 
\end{aligned}
\tag{$\mathbf{M} \rightarrow \mathbf{C} \circlearrowright$}
\label{eq: mixing}
\end{equation}
Here, $\mathrm{Emb}_{\text{[MASK]}}$ denotes the embedding corresponding to the [MASK] token, and $\beta \in (0, 1)$ represents the mixing coefficient between the output distribution and $\mathrm{Emb}_{\text{[MASK]}}$. In addition, when converting the output distribution into an embedding update (\emph{i.e.}, computing $W^{\top} p_{\theta}(\hat{y}_i | X, Y))$, we employ an adaptive top-$p$ sampling strategy that adjusts the threshold by confidence and consolidates residual mass into [MASK] for stability (see full details in~\cref{sec:suppl}).

\begin{algorithm}[!t]
\caption{{\method}}
\label{alg:remix_corrected}
\footnotesize
\begin{algorithmic}[1]
\Require Prompt $X$, Model $p_{\theta}$, Hyperparameters
\State Initialize $S \leftarrow [\mathbf{M}, \ldots, \mathbf{M}]$ (State tracker)
\State Initialize $Y^{\text{emb}}$ with $\mathrm{Emb}_{\text{[MASK]}}$
\State $P_{\text{old}} \leftarrow \mathbf{0}$

\While{$\exists i$ s.t. $S_i \neq \mathbf{T}$}
    \State $P \leftarrow p_{\theta}(\hat{Y} | X, Y^{\text{emb}})$ 
    \Comment{Model forward pass}
    \For{each position $i$ s.t. $S_i \neq \mathbf{T}$}
        \If{Condition  (\cref{eq:decode})}
            \State $S_i \leftarrow \mathbf{T}$
            \State Update $Y^{\text{emb}}_i$  (\cref{eq:decode})
        \Else \Comment{Not decoded}
            \If{$S_i = \mathbf{C}$ \textbf{and} Condition (\cref{eq:reject})}
                \State $S_i \leftarrow \mathbf{M}$ \Comment{Apply $\mathbf{C}\rightarrow\mathbf{M}$}
                \State $Y^{\text{emb}}_i \leftarrow \mathrm{Emb}_{\text{[MASK]}}$
            \Else 
                \State $S_i \leftarrow \mathbf{C}$ \Comment{Apply $\mathbf{M}\rightarrow\mathbf{C}$ or $\mathbf{C}\circlearrowright$}
                \State Update $Y^{\text{emb}}_i$ (\cref{eq: mixing})
            \EndIf
        \EndIf
    \EndFor
    \State $P_{\text{old}} \leftarrow P$ \Comment{Store for next step's \cref{eq:reject}}
\EndWhile
\State \Return Final Tokens from $Y^{\text{emb}}$
\end{algorithmic}
\end{algorithm}

\noindent\textbf{Rejection Rule ($\mathbf{C} \rightarrow \mathbf{M}$).} As a training-free method, although \cref{eq: mixing} mixes output distribution information for the [MASK] tokens at decoding positions, it may introduce inference-training inconsistencies and distributional shifts, since all training is conducted on discrete tokens. To mitigate this issue, we introduce a rejection rule: when a decoding position exhibits significant instability in its output, it is reset to the initial state $\mathrm{M}$. Specifically, when the output distributions of two consecutive decoding steps are too far apart, the position is reset to [MASK].
\begin{equation}
\begin{aligned}
    &y_i \leftarrow \text{[MASK]}, \\ &\ \text{if }\text{position $i$ not decoded, and}\\
    &\ \mathbb{D}_{\text{JS}}(p_{\theta}(\hat{y}_i | X, Y) \parallel p_{\theta}(\hat{y}_i | X, Y_{\text{old}})) > \tau_{\mathrm{rej}},
\end{aligned}
\tag{$\mathbf{C}\rightarrow\mathbf{M}$}
\label{eq:reject}
\end{equation}
where $\mathbb{D}_{\text{JS}}(\cdot \parallel \cdot)$ denotes the Jensen-Shannon (JS) divergence, and $\tau_{\mathrm{rej}}$ is the hyperparameter threshold used to determine instability.

In summary, {\method} orchestrates a dynamic decoding process where positions flexibly transition between masked, continuous, and discrete states. By enabling continuous-space refinement ($\mathbf{M} \rightarrow \mathbf{C} \circlearrowright$) and providing a stability fallback mechanism ($\mathbf{C}\rightarrow\mathbf{M}$), {\method} effectively addresses combinatorial contradictions and accelerates semantic propagation without requiring additional training.

\subsection{Discussion}
The proposed {\method} method offers a novel perspective on stabilizing parallel decoding in DLLMs. By leveraging continuous representations, we bypass the rigid combinatorial constraints imposed by purely discrete sampling. The Mixing Rule can be interpreted as a form of ``soft'' lookahead, allowing undecoded positions to communicate their tentative semantic intent to neighbors without fully committing to a potentially erroneous discrete token. This facilitates a more globally coherent decoding trajectory.

Furthermore, the Rejection Rule acts as a critical regularizer. It implicitly acknowledges the limitations of using continuous states in a model trained for discrete inputs. By resetting unstable positions to [MASK], we effectively prune divergent decoding paths early, preventing error cascading. This training-free nature is a significant advantage, as it avoids the substantial computational cost of re-training or fine-tuning large-scale diffusion models while still addressing the fundamental issue of decoding consistency.

\begin{table*}[!t]
\centering
\caption{Performance and inference speedup comparison on diverse language benchmarks.} \vspace{-10pt}
\label{tab:llada-bench}
\resizebox{\textwidth}{!}{
\begin{tabular}{@{}!{\vrule width 0pt}p{3.5cm}<{\centering}p{1.5cm}<{\centering}p{2.8cm}<{\centering}p{2.8cm}<{\centering}p{2.0cm}<{\centering}p{2.8cm}<{\centering}p{2.0cm}<{\centering}}
\toprule
\multirow{2}{*}{\textbf{Benchmark}} & \multirow{2}{*}{\textbf{Method}} & \multirow{2}{*}{\textbf{Accuracy}} & \multirow{2}{*}{\textbf{Steps}}  & {\textbf{Step}} & \multirow{2}{*}{\textbf{Latency}} & {\textbf{Latency}} \\
&&&& \textbf{Reduction} && \textbf{Speedup}\\
\midrule

\multirow{2}{*}{\shortstack{GSM8K \\ \scriptsize \color{light-gray}{Math Reasoning}}}        & LLaDA & 73.01 & 256 & 1.00 $\times$ & 13.89 & 1.00 $\times$ \\
                              & \cellcolor{ourred} {\method}  & \cellcolor{ourred} 75.66 (\textcolor{ourblue}{+2.65}) & \cellcolor{ourred} 51.55 (\textcolor{ourblue}{-204.45}) & 
                                 \cellcolor{ourred} 4.97 $\times$ & 
                                 \cellcolor{ourred} 3.00 (\textcolor{ourblue}{-10.89}) & 
                                 \cellcolor{ourred} 4.63 $\times$ \\
\midrule
\multirow{2}{*}{\shortstack{MATH-500 \\ \scriptsize \color{light-gray}{Math Reasoning}}}         & LLaDA & 32.20 & 256 & 1.00 $\times$ & 13.96 & 1.00 $\times$ \\
                              & \cellcolor{ourred} {\method}  & \cellcolor{ourred} 35.00 (\textcolor{ourblue}{+2.80}) & \cellcolor{ourred} 66.33 (\textcolor{ourblue}{-189.67}) & 
                                 \cellcolor{ourred} 3.86 $\times$ & 
                                 \cellcolor{ourred} 3.93 (\textcolor{ourblue}{-10.03}) & 
                                 \cellcolor{ourred} 3.55 $\times$ \\
\midrule
\multirow{2}{*}{\shortstack{HumanEval \\ \scriptsize \color{light-gray}{Code Generation}}}   & LLaDA & 43.90 & 256 & 1.00 $\times$ & 19.22 & 1.00 $\times$ \\
                              & \cellcolor{ourred} {\method}  & \cellcolor{ourred} 44.50 (\textcolor{ourblue}{+0.60}) & \cellcolor{ourred} 89.40 (\textcolor{ourblue}{-166.60}) & 
                                 \cellcolor{ourred} 2.86 $\times$ & 
                                 \cellcolor{ourred} 7.13 (\textcolor{ourblue}{-12.09}) & 
                                 \cellcolor{ourred} 2.70 $\times$ \\
\midrule
\multirow{2}{*}{\shortstack{MBPP \\ \scriptsize \color{light-gray}{Code Generation}}}        & LLaDA & 36.20 & 256 & 1.00 $\times$ & 14.11 & 1.00 $\times$ \\
                              & \cellcolor{ourred} {\method}  & \cellcolor{ourred} 37.00 (\textcolor{ourblue}{+0.80}) & \cellcolor{ourred} 84.28 (\textcolor{ourblue}{-171.72}) & 
                                 \cellcolor{ourred} 3.04 $\times$ & 
                                 \cellcolor{ourred} 5.05 (\textcolor{ourblue}{-9.06}) & 
                                 \cellcolor{ourred} 2.79 $\times$ \\
\midrule
\multirow{2}{*}{\shortstack{Countdown \\ \scriptsize \color{light-gray}{Logical Reasoning}}} & LLaDA & 25.78 & 256 & 1.00 $\times$ & 14.39 & 1.00 $\times$ \\
                              & \cellcolor{ourred} {\method}  & \cellcolor{ourred} 29.68 (\textcolor{ourblue}{+3.90}) & \cellcolor{ourred} 101.44 (\textcolor{ourblue}{-154.56}) & 
                                 \cellcolor{ourred} 2.52 $\times$ & 
                                 \cellcolor{ourred} 6.08 (\textcolor{ourblue}{-8.31}) & 
                                 \cellcolor{ourred} 2.37 $\times$ \\
\midrule
\multirow{2}{*}{\shortstack{Sudoku \\ \scriptsize \color{light-gray}{Logical Reasoning}}}    & LLaDA & 14.25 & 256 & 1.00 $\times$ & 19.34 & 1.00 $\times$ \\
                              & \cellcolor{ourred} {\method}  & \cellcolor{ourred} 18.55 (\textcolor{ourblue}{+4.30}) & \cellcolor{ourred} 84.06 (\textcolor{ourblue}{-171.94}) & 
                                 \cellcolor{ourred} 3.05 $\times$ & 
                                 \cellcolor{ourred} 6.76 (\textcolor{ourblue}{-12.58}) & 
                                 \cellcolor{ourred} 2.86 $\times$ \\
\midrule
\multirow{2}{*}{\shortstack{ARC-E \\ \scriptsize \color{light-gray}{Commonsense Reasoning}}}            & LLaDA & 59.68 & 256 & 1.00 $\times$ & 13.70 & 1.00 $\times$ \\
                              & \cellcolor{ourred} {\method}  & \cellcolor{ourred} 70.54 (\textcolor{ourblue}{+10.86}) & \cellcolor{ourred} 50.69 (\textcolor{ourblue}{-205.31}) & 
                                 \cellcolor{ourred} 5.05 $\times$ & 
                                 \cellcolor{ourred} 2.98 (\textcolor{ourblue}{-10.72}) & 
                                 \cellcolor{ourred} 4.60 $\times$ \\
\midrule
\multirow{2}{*}{\shortstack{ARC-C \\ \scriptsize \color{light-gray}{Commonsense Reasoning}}}            & LLaDA & 52.17 & 256 & 1.00 $\times$ & 13.67 & 1.00 $\times$ \\
                              & \cellcolor{ourred} {\method}  & \cellcolor{ourred} 66.22 (\textcolor{ourblue}{+14.05}) & \cellcolor{ourred} 61.30 (\textcolor{ourblue}{-194.70}) & 
                                 \cellcolor{ourred} 4.18 $\times$ & 
                                 \cellcolor{ourred} 3.49 (\textcolor{ourblue}{-10.18}) & 
                                 \cellcolor{ourred} 3.92 $\times$ \\
\bottomrule
\end{tabular}
}
\vspace{-5pt}
\end{table*}

\begin{table*}[t]
\centering
\caption{Performance and inference speedup comparison across diverse multimodal understanding and reasoning benchmarks. We use CIDEr for Flickr30k and accuracy for other benchmarks.}\vspace{-9pt}
\label{tab:mmada-bench}
\resizebox{\textwidth}{!}{
\begin{tabular}{@{}!{\vrule width 0pt}p{3.5cm}<{\centering}p{1.5cm}<{\centering}p{2.8cm}<{\centering}p{2.8cm}<{\centering}p{2.0cm}<{\centering}p{2.8cm}<{\centering}p{2.0cm}<{\centering}}
\toprule
\multirow{2}{*}{\textbf{Benchmark}} & \multirow{2}{*}{\textbf{Method}} & \multirow{2}{*}{\textbf{Performance}} & \multirow{2}{*}{\textbf{Steps}}  & {\textbf{Step}} & \multirow{2}{*}{\textbf{Latency}} & {\textbf{Latency}} \\
&&&& \textbf{Reduction} && \textbf{Speedup}\\
\midrule 
\multirow{2}{*}{\shortstack{Flickr30k-lite \\ \scriptsize \color{light-gray}{Captioning}}}                  & MMaDA & 57.52 & 256 & 1.00 $\times$ & 108.02 & 1.00 $\times$ \\
       & \cellcolor{ourred} {\method}  & \cellcolor{ourred} 59.59 (\textcolor{ourblue}{+2.07}) & \cellcolor{ourred} 30.13 (\textcolor{ourblue}{-225.87}) & 
                                 \cellcolor{ourred} 8.50 $\times$ & 
                                 \cellcolor{ourred} 14.37 (\textcolor{ourblue}{-93.65}) & 
                                 \cellcolor{ourred} 7.52 $\times$ \\
\midrule
\multirow{2}{*}{\shortstack{AI2D-lite \\ \scriptsize \color{light-gray}{Chart Understanding}}}              & MMaDA & 62.40 & 256 & 1.00 $\times$ & 109.27 & 1.00 $\times$  \\
        & \cellcolor{ourred} {\method}  & \cellcolor{ourred} 62.60 (\textcolor{ourblue}{+0.20}) & \cellcolor{ourred} 37.82 (\textcolor{ourblue}{-218.18}) & 
                                 \cellcolor{ourred} 6.77 $\times$ & 
                                 \cellcolor{ourred} 19.24 (\textcolor{ourblue}{-90.03}) & 
                                 \cellcolor{ourred} 5.68 $\times$ \\
\midrule
\multirow{2}{*}{\shortstack{MathVision-mini \\ \scriptsize \color{light-gray}{Math Reasoning}}}            & MMaDA & 12.83  & 256 & 1.00 $\times$ & 111.12 & 1.00 $\times$ \\
        & \cellcolor{ourred} {\method}  & \cellcolor{ourred} 12.83 (\textcolor{ourblue}{+0.00}) & \cellcolor{ourred} 52.71 (\textcolor{ourblue}{-203.29}) & 
                                 \cellcolor{ourred} 4.86 $\times$ & 
                                 \cellcolor{ourred} 27.56 (\textcolor{ourblue}{-83.56}) & 
                                 \cellcolor{ourred} 4.03 $\times$ \\
\midrule
\multirow{2}{*}{\shortstack{MathVista-mini \\ \scriptsize \color{light-gray}{Math Reasoning}}}         & MMaDA & 31.60 & 256 & 1.00 $\times$ & 110.38 & 1.00 $\times$ \\
        & \cellcolor{ourred} {\method} &\cellcolor{ourred} 34.60
                                     (\textcolor{ourblue}{+3.00}) &\cellcolor{ourred} 43.61(\textcolor{ourblue}{-212.39}) &\cellcolor{ourred} 5.87 $\times$ &\cellcolor{ourred} 20.73(\textcolor{ourblue}{-89.65}) &\cellcolor{ourred} 5.32 $\times$ \\
\midrule
\multirow{2}{*}{\shortstack{MMMU-val \\ \scriptsize \color{light-gray}{Multi-discipline Reasoning}}}   & MMaDA & 23.33 & 256 & 1.00 $\times$ & 113.76 & 1.00 $\times$ \\
        & \cellcolor{ourred} {\method}  & \cellcolor{ourred} 24.67 (\textcolor{ourblue}{+1.34}) & \cellcolor{ourred} 58.04 (\textcolor{ourblue}{-197.96}) & 
                                 \cellcolor{ourred} 4.41 $\times$ & 
                                 \cellcolor{ourred} 30.36 (\textcolor{ourblue}{-83.40}) & 
                                 \cellcolor{ourred} 3.75  $\times$ \\
\midrule
\multirow{2}{*}{\shortstack{ScienceQA-IMG \\ \scriptsize \color{light-gray}{Multi-discipline Reasoning}}}  & MMaDA & 46.70 & 256 & 1.00 $\times$ & 111.60 & 1.00 $\times$ \\
        & \cellcolor{ourred} {\method}  & \cellcolor{ourred} 47.35 (\textcolor{ourblue}{+0.65}) & \cellcolor{ourred} 43.86 (\textcolor{ourblue}{-212.14}) & 
                                 \cellcolor{ourred} 5.84 $\times$ & 
                                 \cellcolor{ourred} 21.35 (\textcolor{ourblue}{-69.27}) & 
                                 \cellcolor{ourred} 5.23 $\times$ \\
\bottomrule
\end{tabular}}
\vspace{-10pt}
\end{table*}

\section{Experiments}
\subsection{Experiment setup}
\noindent\textbf{Datasets and Baselines.} We evaluate {\method} across diverse tasks and domains. In the language domain specifically, we compare {\method} with the standard LLaDA decoding on eight tasks:
GSM8K~\citep{DBLP:journals/corr/abs-2110-14168}, MATH-500~\citep{DBLP:conf/nips/HendrycksBKABTS21}, HumanEval~\citep{DBLP:journals/corr/abs-2107-03374}, MBPP~\citep{DBLP:journals/corr/abs-2108-07732}, Countdown~\citep{DBLP:journals/corr/abs-2504-12216}, Sudoku~\citep{DBLP:journals/corr/abs-2504-12216}, ARC-E~\citep{DBLP:journals/corr/abs-1803-05457}, and ARC-C~\citep{DBLP:journals/corr/abs-1803-05457}, encompassing four text-generation categories: mathematical reasoning, code generation, logical reasoning, and commonsense reasoning. For the vision-language domain, we compare {\method} with the standard MMaDA~\citep{DBLP:journals/corr/abs-2505-15809} decoding on six multimodal understanding tasks: Flickr30k~\citep{DBLP:journals/tacl/YoungLHH14}, AI2D~\citep{DBLP:journals/corr/KembhaviSKSHF16}, MathVision~\citep{DBLP:conf/nips/WangPSLRZZL24}, MathVista~\citep{DBLP:conf/iclr/LuBX0LH0CG024}, MMMU~\citep{DBLP:conf/cvpr/YueNZ0LZSJRSWYY24} and ScienceQA~\citep{DBLP:conf/nips/LuMX0CZTCK22}, covering four types of multimodal tasks: captioning, chart understanding, math reasoning and multi-discipline reasoning. For clarity, we run experiments on the lite version of Flickr30k, AI2D from LMMs-Lab~\citep{DBLP:conf/naacl/ZhangLZPCHLZYLL25}, the official testmini subset of MathVision, MathVista, the validation set of MMMU and the ScienceQA-IMG version of ScienceQA.

\noindent\textbf{Evaluation Detail.} All benchmarks are evaluated zero-shot, except Sudoku, which uses a 4-shot setting. We report CIDEr~\cite{DBLP:conf/cvpr/VedantamZP15} for Flickr30k and accuracy for all other benchmarks. For multimodal understanding tasks, we choose GPT-4o mini~\citep{openai2024gpt4omini} as evaluator to assess the correctness and quality of generated answers. To quantify inference efficiency, we measure the required decoding steps and latency(in seconds) for {\method} and baseline on each task, averaging across all samples in the corresponding benchmark.

\noindent\textbf{Implementation Detail.} Our evaluation pipeline is implemented on top of the WINO codebase~\citep{DBLP:journals/corr/abs-2507-18578}, adapting its evaluation scripts to our setting while retaining the default preprocessing and protocol unless otherwise noted. We use LLaDA-8B-Instruct\footnote{\href{https://huggingface.co/GSAI-ML/LLaDA-8B-Instruct}{https://huggingface.co/GSAI-ML/LLaDA-8B-Instruct}} for language evaluation and MMaDA-8B-MixCoT\footnote{\href{https://huggingface.co/Gen-Verse/MMaDA-8B-MixCoT}{https://huggingface.co/Gen-Verse/MMaDA-8B-MixCoT}} for vision–language evaluation. All experiments are conducted on 8 NVIDIA GeForce RTX 3090 GPUs, except for the main language experiments, which are performed on 8 NVIDIA A100 GPUs. We employ semi-autoregressive sampling strategy that partitions the sequence into blocks and generates them left to right. In our experiments, we set the generation length to 256 and block length to 128. For the hyperparameters in {\method}, we set the confidence threshold {$\tau_{\mathrm{conf}}$} to 0.8. The mixing ratio {$\beta$} is selected from \{0.4, 0.5, 0.6\} to balance the trade-off between speed and quality, while the rejection threshold {$\tau_{\mathrm{rej}}$} varies between 0.1 and 0.4.

\begin{table*}[!t]
\centering
\caption{Experiment results on different generation lengths and block lengths}
\vspace{-10pt}
\label{tab:gen_length}
\resizebox{\textwidth}{!}{
\begin{tabular}{@{}!{\vrule width 0pt}p{2.2cm}<{\centering}p{1.8cm}<{\centering}p{1.8cm}<{\centering}p{1.4cm}<{\centering}p{2.8cm}<{\centering}p{1.8cm}<{\centering}p{2.0cm}<{\centering}p{1.8cm}<{\centering}p{1.8cm}<{\centering}}
\toprule
\multirow{2}{*}{\textbf{Benchmark}} & \textbf{Gen} & \textbf{Block} & \multirow{2}{*}{\textbf{Method}} & \multirow{2}{*}{\textbf{Accuracy}} & \multirow{2}{*}{\textbf{Steps}}  & {\textbf{Step}} & \multirow{2}{*}{\textbf{Latency}} & {\textbf{Latency}} \\
& \textbf{Length} & \textbf{Length} & & & & \textbf{Reduction} & & \textbf{Speedup}\\
\midrule\midrule
\multicolumn{9}{c}{\emph{Different Generation Lengths}} \\

\multirow{6}{*}{GSM8K}               & \multirow{2}{*}{128} & \multirow{6}{*}{128} &LLaDA     & 59.21 & 128   & 1.00 $\times$ & 15.15  & 1.00 $\times$        \\
                                     & & & \cellcolor{ourred} {\method} & \cellcolor{ourred} 59.74 (\textcolor{ourblue}{+0.53}) & \cellcolor{ourred} 36.63 &\cellcolor{ourred} 3.49 $\times$ &\cellcolor{ourred} 4.63 &\cellcolor{ourred} 3.27 $\times$      \\
                                     & \multirow{2}{*}{256} &  & LLaDA     & 72.48 & 256   & 1.00 $\times$ & 38.56 & 1.00 $\times$     \\
                                     & & & \cellcolor{ourred} {\method} &\cellcolor{ourred} 75.28 (\textcolor{ourblue}{+2.80}) &\cellcolor{ourred} 58.54 &\cellcolor{ourred} 4.37 $\times$ &\cellcolor{ourred} 8.63 &\cellcolor{ourred} 4.47 $\times$     \\
                                     & \multirow{2}{*}{512} &  & LLaDA     & 74.68 & 512   & 1.00 $\times$ & 114.13 & 1.00 $\times$     \\
                                     &                      & & \cellcolor{ourred} {\method} &\cellcolor{ourred} 77.63 (\textcolor{ourblue}{+2.95}) &\cellcolor{ourred} 74.76 &\cellcolor{ourred} 6.85 $\times$ &\cellcolor{ourred} 17.66 &\cellcolor{ourred} 6.46 $\times$     \\
\midrule
\multirow{6}{*}{MathVista-mini} & \multirow{2}{*}{128} & \multirow{6}{*}{128} & MMaDA     & 31.80 & 128   & 1.00 $\times$ & 44.60  & 1.00 $\times$ \\
                                     & & &\cellcolor{ourred} {\method} &\cellcolor{ourred} 33.80 (\textcolor{ourblue}{+2.00}) &\cellcolor{ourred} 35.87 &\cellcolor{ourred} 3.57 $\times$ &\cellcolor{ourred} 16.15 &\cellcolor{ourred} 2.76 $\times$ \\
                                     & \multirow{2}{*}{256} & & MMaDA     & 31.60 & 256 & 1.00 $\times$ & 110.38 & 1.00 $\times$     \\
                                     &                      & &\cellcolor{ourred} {\method} &\cellcolor{ourred} 34.60
                                     (\textcolor{ourblue}{+3.00}) &\cellcolor{ourred} 43.61 &\cellcolor{ourred} 5.87 $\times$ &\cellcolor{ourred} 20.73 &\cellcolor{ourred} 5.32 $\times$  \\
                                     & \multirow{2}{*}{512} & & MMaDA     & 31.30 & 512 & 1.00 $\times$ & 242.9 & 1.00 $\times$     \\
                                     &                      & &\cellcolor{ourred} {\method} &\cellcolor{ourred} 33.40 
                                     (\textcolor{ourblue}{+2.10}) &\cellcolor{ourred} 71.31 &\cellcolor{ourred} 7.18 $\times$ &\cellcolor{ourred} 38.81 &\cellcolor{ourred} 6.26 $\times$  \\
\midrule\midrule
\multicolumn{9}{c}{\emph{Different Block Lengths}} \\
\multirow{6}{*}{GSM8K}               & \multirow{6}{*}{256} & \multirow{2}{*}{32} &LLaDA     & 81.73 & 256   & 1.00 $\times$ & 35.32  & 1.00 $\times$        \\
                                     & & & \cellcolor{ourred} {\method} & \cellcolor{ourred} 81.88 (\textcolor{ourblue}{+0.15}) & \cellcolor{ourred} 66.43 &\cellcolor{ourred} 3.85 $\times$ &\cellcolor{ourred} 9.66 &\cellcolor{ourred} 3.66 $\times$      \\
                                     & & \multirow{2}{*}{64} & LLaDA     & 82.56 & 256   & 1.00 $\times$ & 38.77 & 1.00 $\times$     \\
                                     & & & \cellcolor{ourred} {\method} &\cellcolor{ourred} 82.48 (\textcolor{ourblue}{-0.08}) &\cellcolor{ourred} 63.04 &\cellcolor{ourred} 4.06 $\times$ &\cellcolor{ourred} 9.49 &\cellcolor{ourred} 4.09 $\times$     \\
                                     &  & \multirow{2}{*}{128} & LLaDA     & 72.48 & 256   & 1.00 $\times$ & 38.56 & 1.00 $\times$     \\
                                     & & & \cellcolor{ourred} {\method} &\cellcolor{ourred} 75.28 (\textcolor{ourblue}{+2.80}) &\cellcolor{ourred} 58.54 &\cellcolor{ourred} 4.37 $\times$ &\cellcolor{ourred} 8.63 &\cellcolor{ourred} 4.47 $\times$  \\
\midrule
\multirow{6}{*}{MathVista-mini} & \multirow{6}{*}{256} & \multirow{2}{*}{32} & MMaDA     & 33.00 & 256 & 1.00 $\times$ & 109.39  & 1.00 $\times$ \\
                                     & & &\cellcolor{ourred} {\method} &\cellcolor{ourred} 34.40 (\textcolor{ourblue}{+1.40}) &\cellcolor{ourred} 55.08 &\cellcolor{ourred} 4.65 $\times$ &\cellcolor{ourred} 25.11 &\cellcolor{ourred} 4.36 $\times$ \\
                                     & & \multirow{2}{*}{64} & MMaDA     & 31.70 & 256 & 1.00 $\times$ & 109.84 & 1.00 $\times$     \\
                                     &                      & &\cellcolor{ourred} {\method} &\cellcolor{ourred} 33.90
                                     (\textcolor{ourblue}{+2.20}) &\cellcolor{ourred} 54.49 &\cellcolor{ourred} 4.70 $\times$ &\cellcolor{ourred} 25.69 &\cellcolor{ourred} 4.28 $\times$  \\
                                     & & \multirow{2}{*}{128} & MMaDA & 31.60 & 256 & 1.00 $\times$ & 110.38 & 1.00 $\times$     \\
                                     &                      & &\cellcolor{ourred} {\method} &\cellcolor{ourred} 34.60
                                     (\textcolor{ourblue}{+3.00}) &\cellcolor{ourred} 43.61 &\cellcolor{ourred} 5.87 $\times$ &\cellcolor{ourred} 20.73 &\cellcolor{ourred} 5.32 $\times$  \\
\bottomrule
\end{tabular}
}\vspace{-10pt}
\end{table*}

\subsection{Main Results}
\noindent\textbf{Performance gain and Speedup in the Language Domain.} We evaluate the performance and efficiency of LLaDA with and without {\method} across eight language benchmarks, as shown in \cref{tab:llada-bench}.
Across eight diverse reasoning and coding benchmarks, incorporating {\method} consistently 
improves both accuracy and efficiency over LLaDA, demonstrating that our method delivers generalizable gains rather than task-specific spikes. On the accuracy side, {\method} yields improvements on every benchmark, ranging from modest boosts in code generation tasks (e.g., +0.60 on HumanEval, +0.80 on MBPP) to substantial jumps on challenging reasoning datasets (e.g., +3.90 on Countdown, +4.30 on Sudoku, and up to +14.05 on ARC-C).
Notably, even in math reasoning tasks where LLaDA is relatively strong, 
{\method} still achieves significant improvements (+2.65 on GSM8K, +2.80 on MATH-500). 

Efficiency gains are equally consistent. {\method} reduces the number of sampling steps by 2.5$\times$–5.0$\times$, corresponding to absolute reductions of 150–205 steps compared to the 256-step LLaDA baseline. This directly translates into substantially lower generation latency: latency drops by 8–13 seconds, producing 2.4$\times$–4.6$\times$ end-to-end speedup depending on the dataset. Importantly, these efficiency improvements do not come at the cost of accuracy—in all cases, accuracy increases alongside faster inference.

\noindent\textbf{Performance Gain and Speedup in the Multimodal Domain.} We further evaluate {\method} in the multimodal setting by integrating it into the MMADA framework and conducting experiments on six vision–language benchmarks, as reported in \cref{tab:mmada-bench}.
{\method} again shows consistent improvements over MMaDA, confirming that the benefits of our approach extend beyond text-only reasoning into the multimodal domain.
On performance, Remix delivers accuracy gains on nearly all datasets, with improvements ranging from +0.20 on AI2D-lite to +3.00 on MathVista-mini, and notably +2.07 on Flickr30k-lite. Efficiency improvements are substantial across all benchmarks. Remix reduces sampling steps by 4.4$\times$ to 8.5$\times$. As a result, inference becomes 3.75$\times$ to 7.52$\times$ faster, saving ~70–94 seconds per query depending on the task. In particular, captioning and chart-understanding benchmarks achieve the largest speedups (7.5$\times$ and 5.7$\times$).

\begin{figure}[!t]
    \centering
    \includegraphics[width=\linewidth]{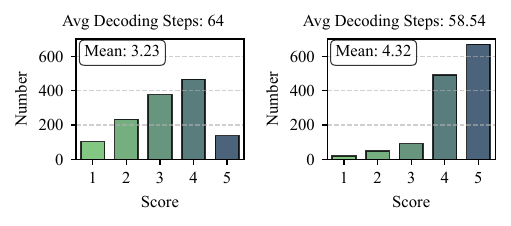}
    \vspace{-24pt}
    \caption{Comparison of GSM8K responses~(LLaDA-based) score distributions addressing combinatorial contradictions on naive parallel decoding~(left) and {\method}~(right)}
    \label{fig:score_distributions}
    \vspace{-20pt}
\end{figure}

\begin{figure*}[t]
  \centering
  \begin{minipage}[t]{0.49\textwidth}
    \centering
    \includegraphics[width=\linewidth]{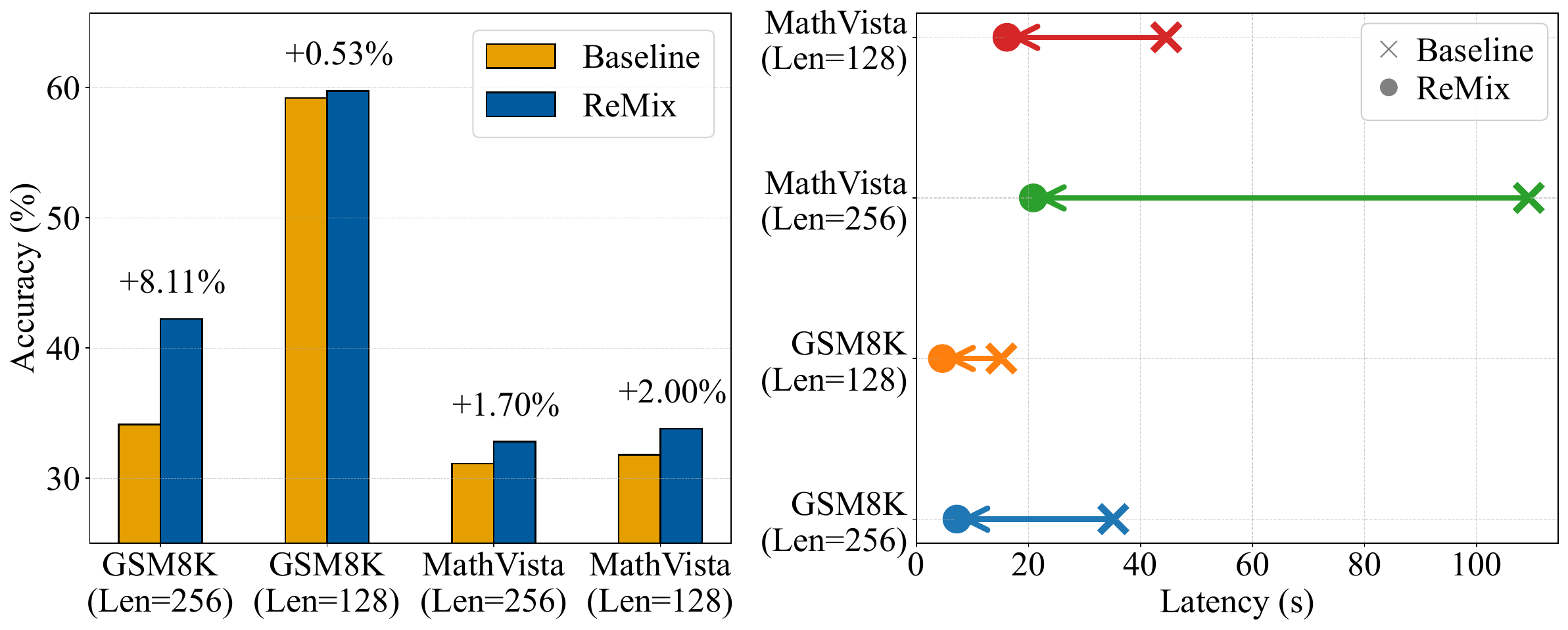}
    \vspace{-17pt}
    \captionof{figure}{Experiment results on fully diffusion-based decoding~(generation length = block length).}
    \label{fig:full_diffusion}
  \end{minipage}\hfill
  \begin{minipage}[t]{0.49\textwidth}
    \centering
    \includegraphics[width=0.49\linewidth]{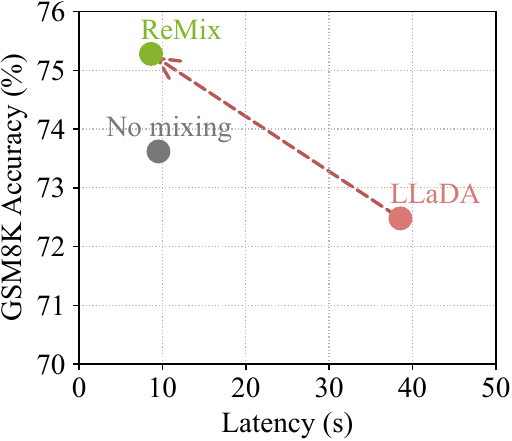}
    \includegraphics[width=0.49\linewidth]{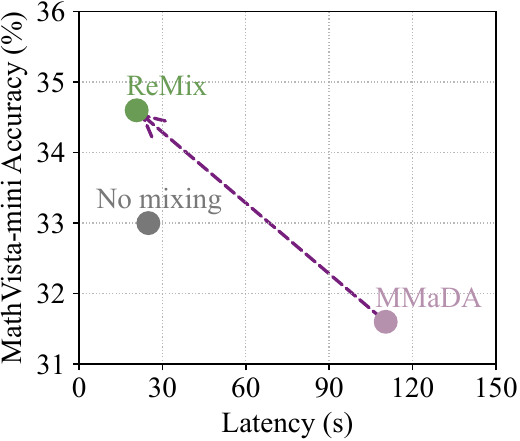}
    \vspace{-5pt}
    \captionof{figure}{Ablation of the mixing module on GSM8K~(left) and MathVista-mini~(right).}
    \label{fig:ablation_mixing}
  \end{minipage}
  \vspace{-10pt}
\end{figure*}
\begin{figure*}[!t]
    \centering
    \begin{minipage}{0.49\textwidth} 
        \centering
        \includegraphics[width=\linewidth]{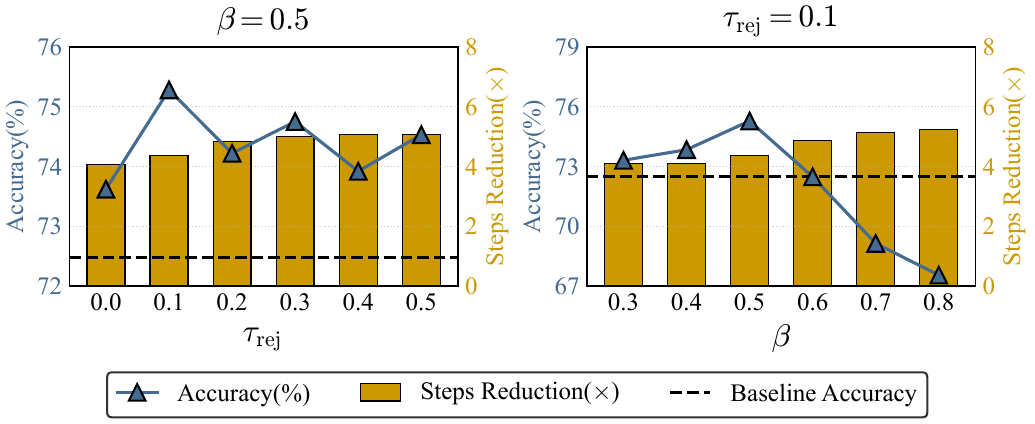}
        \subcaption{{\method}~(LLaDA-based) on GSM8K} 
        \label{fig:ablation_llada}
    \end{minipage}
    \hfill 
    \begin{minipage}{0.49\textwidth}
        \centering
        \includegraphics[width=\linewidth]{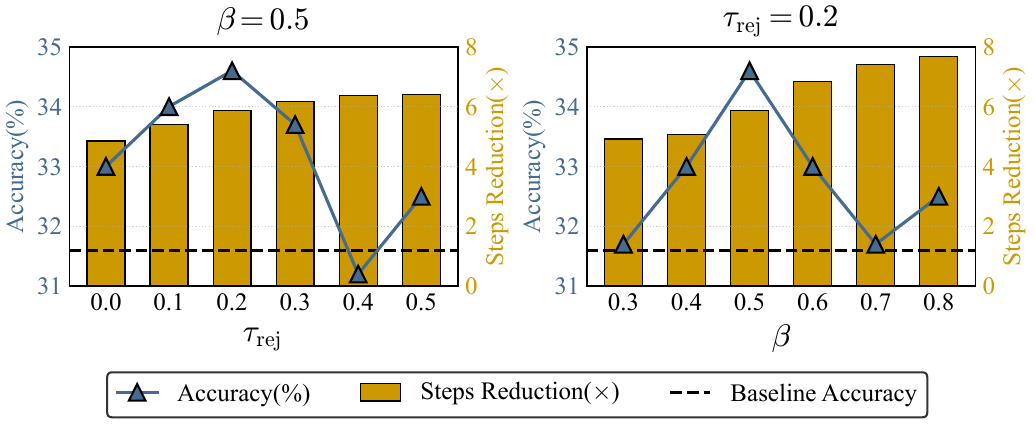} 
        \subcaption{{\method}~(MMaDA-based) on MathVista-mini} 
        \label{fig:ablation_mmada} 
    \end{minipage}
    \vspace{-8pt}
    \caption{Ablation study on the mixing ratio {$\beta$} and rejection threshold {$\tau_{rej}$}} 
    \label{fig:ablation_comparison} 
    \vspace{-10pt}
\end{figure*}

\subsection{Ablation Study and Further Analysis}

\noindent\textbf{Performance Across Different Generation and Block Lengths.} In the first half of \cref{tab:gen_length}, we evaluate the performance of {\method} across various generation lengths, keeping the block length fixed to 128. {\method} consistently shows a significant imp rovement in accuracy for all generation lengths tested, while also substantially reducing the number of decoding steps. Remarkably, the performance gains are more pronounced with longer generation lengths. For instance, when the generation length is set to 512, {\method} achieves accuracy improvements of 2.95\% and 2.10\% on GSM8K and MathVista-mini, while accelerating decoding efficiency by 6.46$\times$ and 6.26$\times$, respectively. 
Similarly, we present the results of {\method} with different block lengths in the lower half of \cref{tab:gen_length}, where the generation length is fixed to 256. In this case, {\method} consistently delivers comparable or better accuracy across all three block length settings, accompanied by significant reductions in inference time.
In summary, {\method} proves robustness w.r.t. both generation length and block length. It maintains high accuracy while consistently achieving stable acceleration in model inference, regardless of variations in these parameters.

\noindent\textbf{Analysis on the Mitigation of Combinatorial Contradictions.} 
To demonstrate the effectiveness of {\method} in mitigating ``combinatorial contradiction", we use GPT-4o-mini to evaluate the generated responses on the GSM8K language benchmark, scoring them from 1 to 5, with lower scores indicating poorer performance. As shown in \cref{fig:score_distributions}, our method achieves a significantly higher score of 4.32 compared to 3.23 from the baseline, with fewer decoding steps, and approaches the score of 4.33 achieved by standard decoding. The distribution further highlights the improvement: in the baseline, samples are mainly distributed between scores of 2 and 4, with fewer than 200 samples rated 5. In contrast, {\method} results in a distribution skewed towards scores of 4 and 5, with over 600 samples receiving a perfect score of 5.
The significant improvement in performance, as shown by both the higher average score and the better distribution of results, demonstrates the effectiveness of {\method} in alleviating the degradation caused by the combinatorial contradiction during parallel decoding.

\noindent\textbf{Performance on Fully Diffusion-Based Decoding} In \cref{fig:full_diffusion}, we examine the full-diffusion scenario, where the generation length matches the block length. The results demonstrate that {\method} consistently achieves both accuracy improvements and speedup on both baselines. It's worth mentioning that LLaDA baseline(Len=256) experiences severe performance degradation compared to the semi-autoregressive decoding paradigm shown in \cref{tab:llada-bench}, whereas {\method} compensates for this loss with far fewer decoding steps. This highlights that our method effectively enhances baseline full-diffusion decoding by incorporating continuous representations.

\noindent\textbf{Ablation on Mixing Module.} We perform an ablation study on the mixing module of our method by setting the mixing ratio to 0, thereby disabling the intermediate continuous state $\mathbf{C}$. This results in a degeneration of our method to confidence-aware parallel decoding strategy~\citep{DBLP:journals/corr/abs-2505-22618}. As illustrated in \cref{fig:ablation_mixing}, our approach outperforms this variant, achieving significantly higher accuracy while maintaining decoding speedup. These findings provide compelling evidence that the incorporation of continuous embeddings enhances the model generation capability.

\noindent\textbf{Tuning of Mixing Ratio and Rejection Threshold.} In \cref{fig:ablation_comparison}, we present the results of {\method} with varying mixing ratio $\beta$ and rejection threshold $\tau_{\mathrm{rej}}$, two key parameters that govern the speed-quality trade-off of the method. 
For both GSM8K and MathVista-mini, accuracy initially increases as $\beta$ grows. A moderate level of mixing strengthens the contribution of the continuous embedding, enabling the model to capture richer contextual signals and improving solution quality. However, when $\beta$ becomes too large (e.g., $\beta>0.6$ in GSM8K or $\beta>0.5$ in MathVista), performance begins to degrade sharply, which reflects a shift toward over-reliance on continuous updates, destabilizing the hybrid decoding process.
A similar pattern emerges with $\tau_{\mathrm{rej}}$. Increasing the threshold initially helps accuracy by allowing the model to accept more continuous updates when they are confident, improving the refinement of intermediate predictions. Beyond a certain point, however, overly permissive thresholds introduce low-confidence or noisy continuous embeddings, which in turn reduce accuracy. Across all tested values, step reduction remains large and stable, and {\method} surpasses the baseline in nearly all configurations, demonstrating resilience to a wide range of $\tau_{\mathrm{rej}}$ values.

\section{Conclusion}
This work identifies combinatorial contradiction as a fundamental obstacle limiting the quality–speed trade-off in parallel DLLM decoding, which arises due to  the rigidity of purely discrete updates that fail to preserve cross-position dependencies when many tokens are decoded simultaneously. 
We thus introduce \method, a simple, training-free decoding framework that augments masked discrete diffusion with a continuous mixing state and a lightweight rejection mechanism. {\method} consistently delivers 2–8$\times$ inference acceleration without sacrificing output quality and often improves it, narrowing the performance gap with AR models while retaining DLLMs’ advantages. These results highlight the promise of integrating  continuous and discrete diffusion and point toward a new family of efficient generative models.

\section{Acknowledgment}
This work is supported by National Natural Science Foundation of China (No.~62306178) and STCSM (No.~22DZ2229005).

{
    \small
    \bibliographystyle{ieeenat_fullname}
    \bibliography{main}
}

\clearpage
\setcounter{page}{1}
\appendix  %
\onecolumn %
\maketitlesupplementary %

\section{Adaptive Top-$p$ Sampling in Mixing Rule}\label{sec:suppl}
As mentioned in~\cref{sec:method}, we employ top-$p$ (nucleus) sampling during the embedding update to improve inference stability. 
Rather than using a fixed nucleus threshold $p$ across all positions, we adopt a dynamic strategy that adapts to the model's prediction confidence at each position. 
Specifically, let $p_{\max}$ denote the highest probability in the output distribution at the current position. We define the adaptive nucleus threshold as:
\begin{equation}
    p_{\mathrm{dyn}} = \min\left(2 \cdot p_{\max},\; 0.9\right).
    \nonumber
\end{equation}
The residual probability mass, i.e., the portion not included in this top set, is assigned to the [MASK] token. To enhance numerical and behavioral stability during iterative refinement, this design prevents low-confidence, low-probability tokens from contributing to the embedding update; instead, their weight is consolidated into the [MASK] embedding, which serves as a stable anchor in the continuous state $\mathbf{C}$.

\section{Additional Experiments}

\subsection{Comparison with State-of-the-Art Methods}
We compare {\method} with three SOTA accelerating methods on GSM8K: WINO, Fast-dLLM, and Learn2PD. 
WINO and {\method} are evaluated under default settings, while Fast-dLLM and Learn2PD follow their optimal configuration with block length 32 (Learn2PD uses the official adapter\footnote{\href{https://github.com/ims-kdks/Learning-to-Parallel-Decoding}{https://github.com/ims-kdks/Learning-to-Parallel-Decoding}}). 
As shown in \cref{tab:sota_results}, {\method} achieves the best speed-quality trade-off, outperforming baselines in throughput without sacrificing accuracy.

\begin{table}[h]
\centering
\caption{Accuracy and throughput (TPS) comparison on GSM8K.}
\vspace{-10pt}
\label{tab:sota_results}
\begin{tabular}{lcccc}
\hline
\multirow{2}{*}{\textbf{Method}} 
& \multicolumn{2}{c}{\textbf{Block Length = 128}} 
& \multicolumn{2}{c}{\textbf{Block Length = 32}} \\
\cline{2-5}
& \textbf{Acc.} & \textbf{TPS} 
& \textbf{Acc.} & \textbf{TPS} \\
\hline
LLaDA      & 72.55 & 22.15 & 81.88 & 22.20 \\
WINO       & 75.28\textcolor{blue}{(+2.73)} & 74.82\textcolor{red}{($\times$3.38)} & -- & -- \\
Fast-dLLM  & -- & -- & 80.29\textcolor{blue}{(-1.59)} & 75.45\textcolor{red}{($\times$3.40)} \\
Learn2PD   & -- & -- & 78.17\textcolor{blue}{(-3.71)} & 102.72\textcolor{red}{($\times$4.63)} \\
ReMix      & 75.36\textcolor{blue}{(+2.81)} & 102.47\textcolor{red}{($\times$4.63)} & 81.20\textcolor{blue}{(-0.68)} & 80.29\textcolor{red}{($\times$3.62)} \\
\hline
\end{tabular}
\vspace{-12pt}
\end{table}

\subsection{Analysis of Computational Overhead} 
We profile the average overhead on GSM8K in \cref{tab:overhead}, where ReMix-specific computations account for only 9.12\% of the overall runtime. This efficiency stems from the lightweight design of both modules: the Mixing update contains highly parallelizable projection restricted to undecoded positions, while the Rejection rule involves only efficient distribution-wise divergence calculations. Since neither operation requires extra backbone forward passes, the overhead remains negligible compared to the dominant transformer computation.

\begin{table}[h]
\centering
\caption{Runtime breakdown of ReMix overhead on GSM8K.}
\vspace{-10pt}
\label{tab:overhead}
\begin{tabular}{lrr}
\hline
\textbf{Component} & \textbf{Time (s)} & \textbf{Share (\%)} \\
\hline
End-to-end generation & 2.52 & 100.00 \\
\hline
Mixing update & 0.14 & 5.56 \\
Rejection & 0.09 & 3.56 \\
\hline
ReMix-specific total & 0.23 & 9.12 \\
\hline
\end{tabular}
\vspace{-12pt}
\end{table} 

\section{Case Study}
To facilitate a comprehensive comparison of the decoding processes between the baseline methods and {\method}, we present two generation examples: one from GSM8K and another from MMMU-val. As demonstrated, the standard decoding procedure can prematurely predict incorrect tokens, which significantly disrupts the generation of subsequent tokens. In contrast, {\method} effectively mitigates this error propagation by incorporating continuous information into the decoding process, thereby preventing such errors from occurring in the first place.

\begin{tcolorbox}[title={Case: GSM8K, Generation Length=256, Block Length=128}, breakable]
\noindent\textbf{Question:} Randy has 9 oatmeal cookies, 4 chocolate chip cookies, and 5 sugar cookies. He ate 3 cookies for an early day snack, one of each flavor. He ate 2 oatmeal cookies for lunch. He gives 2 sugar cookies to his friends. Then, he bakes 4 of each flavor for dinner. How many cookies does he have now?

\noindent\textbf{GT:} 23

\noindent\rule{\linewidth}{0.5pt}

\noindent\textbf{LLaDA:} 

\noindent\hdashrule{\linewidth}{0.5pt}{2pt 2pt}

{\bf \color{gray} Step: 239 (\textcolor{red}{Decoding 2nd block, error occurs}):}

First, let's determine the initial number of each type of cookie:

- Oatmeal cookies: 9

- Chocolate chip cookies: 4

- Sugar cookies: 5

Next, we account for the cookies Randy ate and gave away:

- He ate 3 cookies for an early day snack, one of each flavor:
  
  - Oatmeal cookies: 9 - 1 = 8
  
  - Chocolate chip cookies: 4 - 1 = 3
  
  - Sugar cookies: 5 - 1 = 4

- He ate 2 oatmeal cookies for lunch:
  
  - Oatmeal cookies: 8 - 2 = 6

- He gave 2 sugar cookies to his friends:
 
  - Sugar cookies: 4 - 2 = 2

, he bakes 4 of each flavor for dinner:

- Oatmeal cookies: 6 + 4 = 10

- Chocolate chip cookies: 3

- Sugar cookies: 2 + 4 = 6

\texttt{<}/reasoning\texttt{>}

\texttt{<}answer\texttt{>}

\boxed{\textcolor{red}{1}}

\noindent\hdashrule{\linewidth}{0.5pt}{2pt 2pt}

{\bf \color{gray} Step: 239 (\textcolor{red}{Decoding 2nd block, error accumulates}):}

First, let's determine the initial number of each type of cookie:

- Oatmeal cookies: 9

- Chocolate chip cookies: 4

- Sugar cookies: 5

Next, we account for the cookies Randy ate and gave away:

- He ate 3 cookies for an early day snack, one of each flavor:

  - Oatmeal cookies: 9 - 1 = 8
  
  - Chocolate chip cookies: 4 - 1 = 3
  
  - Sugar cookies: 5 - 1 = 4

- He ate 2 oatmeal cookies for lunch:

  - Oatmeal cookies: 8 - 2 = 6

- He gave 2 sugar cookies to his friends:

  - Sugar cookies: 4 - 2 = 2

, he bakes 4 of each flavor for dinner:

- Oatmeal cookies: 6 + 4 = 10

- Chocolate chip cookies: 3

- Sugar cookies: 2 + 4 = 6

\texttt{<}/reasoning\texttt{>}

\texttt{<}answer\texttt{>}

\boxed{\textcolor{red}{19}}

\noindent\hdashrule{\linewidth}{0.5pt}{2pt 2pt}

{\bf \color{gray} Step: 255 (\textcolor{red}{Final result}):}

First, let's determine the initial number of each type of cookie:

- Oatmeal cookies: 9

- Chocolate chip cookies: 4

- Sugar cookies: 5

Next, we account for the cookies Randy ate and gave away:

- He ate 3 cookies for an early day snack, one of each flavor:
  
  - Oatmeal cookies: 9 - 1 = 8
  
  - Chocolate chip cookies: 4 - 1 = 3
  
  - Sugar cookies: 5 - 1 = 4

- He ate 2 oatmeal cookies for lunch:
  
  - Oatmeal cookies: 8 - 2 = 6

- He gave 2 sugar cookies to his friends:
  
  - Sugar cookies: 4 - 2 = 2

Finally, he bakes 4 of each flavor for dinner:

- Oatmeal cookies: 6 + 4 = 10

- Chocolate chip cookies: \textcolor{red}{3}

- Sugar cookies: 2 + 4 = 6

Total cookies: \textcolor{red}{10 + 3 + 6 = 1}

\texttt{<}/reasoning\texttt{>}

\texttt{<}answer\texttt{>}

\boxed{\textcolor{red}{19}}

\noindent\rule{\linewidth}{0.5pt}

\textbf{\method:}

\noindent\hdashrule{\linewidth}{0.5pt}{2pt 2pt}

{\bf \color{gray} Step: 47 (\textcolor{red}{Decoding 2nd block, no error}):}

First, let's determine the initial number of each type of cookie:

- Oatmeal cookies: 9

- Chocolate chip cookies: 4

- Sugar cookies: 5

Next, we account for the cookies Randy ate:

- He ate 3 cookies for an early day snack, one of each flavor:
  
  Oatmeal cookies: 9 - 1 = 8
  
  Chocolate chip cookies: 4 - 1 = 3
  
  Sugar cookies: 5 - 1 = 4

- He ate 2 oatmeal cookies for lunch:
  
  Oatmeal cookies: 8 - 2 = 6

- He 2 sugar cookies to his friends:
  
  Sugar cookies: 4 - 2 = 2

, he bakes 4 of each flavor for dinner:

- Oatmeal cookies: 6 + 4 = 10

- Chocolate chip cookies: 3 cookies

\texttt{<}/reasoning\texttt{>}

\texttt{<}answer\texttt{>}

\boxed{\textcolor{red}{2}}

\noindent\hdashrule{\linewidth}{0.5pt}{2pt 2pt}

{\bf \color{gray} Step: 48 (\textcolor{red}{Decoding 2nd block, correct induction}):}

First, let's determine the initial number of each type of cookie:

- Oatmeal cookies: 9

- Chocolate chip cookies: 4

- Sugar cookies: 5

Next, we account for the cookies Randy ate:

- He ate 3 cookies for an early day snack, one of each flavor:
  
  Oatmeal cookies: 9 - 1 = 8
 
  Chocolate chip cookies: 4 - 1 = 3
  
  Sugar cookies: 5 - 1 = 4

- He ate 2 oatmeal cookies for lunch:
  
  Oatmeal cookies: 8 - 2 = 6

- He 2 sugar cookies to his friends:
  
  Sugar cookies: 4 - 2 = 2

, he bakes 4 of each flavor for dinner:

- Oatmeal cookies: 6 + 4 = 10

- Chocolate chip cookies: \textcolor{red}{3 + 4 = 7}

- Sugar cookies: 2 + 4 = 

\texttt{<}/reasoning\texttt{>}

\texttt{<}answer\texttt{>}

\boxed{\textcolor{red}{2}}

\noindent\hdashrule{\linewidth}{0.5pt}{2pt 2pt}

{\bf \color{gray} Step: 64 (\textcolor{red}{Final result}):}

First, let's determine the initial number of each type of cookie:

- Oatmeal cookies: 9

- Chocolate chip cookies: 4

- Sugar cookies: 5

Next, we account for the cookies Randy ate:

- He ate 3 cookies for an early day snack, one of each flavor:
 
  Oatmeal cookies: 9 - 1 = 8
  
  Chocolate chip cookies: 4 - 1 = 3
  
  Sugar cookies: 5 - 1 = 4

- He ate 2 oatmeal cookies for lunch:
  
  Oatmeal cookies: 8 - 2 = 6

- He gave 2 sugar cookies to his friends:
  
  Sugar cookies: 4 - 2 = 2

Then, he bakes 4 of each flavor for dinner:

- Oatmeal cookies: 6 + 4 = 10

- Chocolate chip cookies: \textcolor{red}{3 + 4 = 7}

- Sugar cookies: 2 + 4 = 6

Adding these together:

Total cookies = \textcolor{red}{10 + 7 + 6 = 23} cookies

\texttt{<}/reasoning\texttt{>}

\texttt{<}answer\texttt{>}

\boxed{\textcolor{red}{23}}

\end{tcolorbox}

\begin{tcolorbox}[title={Case: MMMU-val, Generation Length=256, Block Length=128}, breakable]

\noindent\textbf{Question:} Which statement is false? 

(A) The differential diagnosis on this case is remote infarct vs. remote contusion

(B) There is a well developed gliotic wall associated with this remote injury

(C) The presence of macrophages is not helpful in distinguishing this from a subacute process

(D) This is very unlikely to have resulted from venous occlusion

\noindent\textbf{Image:}

\vspace{\medskipamount}
{\includegraphics[width=0.3\textwidth]{fig/336.jpg}}
\vspace{\medskipamount}

\noindent\textbf{GT:} A

\noindent\rule{\linewidth}{0.5pt}

\noindent\textbf{MMaDA:} 

\noindent\hdashrule{\linewidth}{0.5pt}{2pt 2pt}

{\bf \color{gray} Step: 119 (\textcolor{red}{Decoding 1st block, error occurs}):}

\texttt{<}think\texttt{>}

To determine which statement is false, let\'s analyze each option:

1. "The differential diagnosis on this case is remote infarct vs. remote contusion."

- This is because the image clearly shows a distant infarct a.

2. "There is a well developed gliotic wall associated with this remote injury."

- This is because the image shows a well gliotic wall, which is in the.

3. "The presence of macrophages is not helpful in distinguishing this from a subacute process."

- This is \textcolor{red}{false}

\noindent\hdashrule{\linewidth}{0.5pt}{2pt 2pt}

{\bf \color{gray} Step: 121 (\textcolor{red}{Decoding 1st block, error accumulates}):}

\texttt{<}think\texttt{>}

To determine which statement is false, let\'s analyze each option:

1. "The differential diagnosis on this case is remote infarct vs. remote contusion."

- This is \textcolor{red}{true} because the image clearly shows a distant infarct a.

2. "There is a well developed gliotic wall associated with this remote injury."

- This is \textcolor{red}{true} because the image shows a well gliotic wall, which is in the.

3. "The presence of macrophages is not helpful in distinguishing this from a subacute process."

- This is \textcolor{red}{false}

\noindent\hdashrule{\linewidth}{0.5pt}{2pt 2pt}

{\bf \color{gray} Step: 256 (\textcolor{red}{Final result}):}

\texttt{<}think\texttt{>}

To determine which statement is false, let\'s analyze each option:

1. ``The differential diagnosis on this case is remote infarct vs. remote contusion."

- This is \textcolor{red}{true} because the image clearly shows a distant infarct from a distant location.

2. ``There is a well developed gliotic wall associated with this remote injury."

- This is \textcolor{red}{true} because the image shows a well developed gliotic wall, which is visible in the foreground.

3. ``The presence of macrophages is not helpful in distinguishing this from a subacute process."

- This is \textcolor{red}{false} because the image shows a subacute process with macrophages visible.

Therefore, the false statement is \textcolor{red}{C}.

\texttt{<}/think\texttt{>}

\textcolor{red}{C}

\noindent\rule{\linewidth}{0.5pt}

\noindent\textbf{ReMix:}

\noindent\hdashrule{\linewidth}{0.5pt}{2pt 2pt}

{\bf \color{gray} Step: 12 (\textcolor{red}{Decoding 1st block, no error}):}

To determine which statement is false, let\'s analyze each option:

1. "The differential diagnosis on this case is vs. remote contusion."

This is \textcolor{red}{false} because the image clearly shows a.

2. "There is a well developed gliotic wall associated with this remote injury."

This is because a well gliotic wall is a feature in the image.

3. "The presence of macrophages is not helpful in distinguishing this from a subacute process."

This is'

\noindent\hdashrule{\linewidth}{0.5pt}{2pt 2pt}

{\bf \color{gray} Step: 47 (\textcolor{red}{Final result}):}

\texttt{<}think\texttt{>}

To determine which statement is false, let\'s analyze each option:

1. ``The differential diagnosis on this case is remote injury vs. is remote contusion."

This is \textcolor{red}{false} because the image clearly shows a distant regional contusion, not a distant infarct.

2. ``There is a well developed gliotic wall associated with this remote injury."

This is \textcolor{red}{true} because a well developed gliotic wall is a clear feature in the image.

3. ``The presence of macrophages is not helpful in distinguishing this from a subacute process."

This is \textcolor{red}{true} because the presence of macrophages in subacute processes is a characteristic feature of subacute processes.

4. ``This is very unlikely to have resulted from venous occlusion."

This is \textcolor{red}{true} because the image clearly shows a distant infarct, not venous occlusion.

Therefore, the false statement is option \textcolor{red}{A}.

\texttt{<}/think\texttt{>}

\textcolor{red}{A}

\end{tcolorbox}

\end{document}